\documentclass[fleqn,10pt]{wlscirep}
\usepackage[utf8]{inputenc}
\usepackage[T1]{fontenc}
\usepackage{bm}
\usepackage{graphicx}
\usepackage{sidecap}
\usepackage{subcaption}
\usepackage{mwe}
\usepackage{wrapfig}
\usepackage{float}
\usepackage{caption}
\usepackage{xcolor}
\usepackage{soul}
\usepackage{amsmath}
\usepackage{gensymb}

\title{Multidimensional analysis using sensor arrays with deep learning for high-precision and high-accuracy diagnosis}

\author[1]{Julie Payette}
\author[1,*]{Sylvain Cloutier}
\author[1]{Fabrice Vaussenat}

\affil[1]{École de technologie supérieure, Department of Electrical Engineering, Montréal, H3C 1K3, Canada}

\affil[*]{SylvainG.Cloutier@etsmtl.ca}

\begin{abstract}
In the upcoming years, artificial intelligence (AI) is going to transform the practice of medicine in most of its specialties. Deep learning can help achieve better and earlier problem detection, while reducing errors on diagnosis. By feeding a deep neural network (DNN) with the data from a low-cost and low-accuracy sensor array, we demonstrate that it becomes possible to significantly improve  the measurements' precision and accuracy. 
The data collection is done with an array composed of 32 temperature sensors, including 16 analog and 16 digital sensors. All sensors have accuracies between 0.5-2.0$ \degree$C. 800 vectors are extracted, covering a range from to 30 to 45$ \degree$C. In order to improve the temperature readings, we use machine learning to perform a linear regression analysis through a DNN. In an attempt to minimize the model's complexity in order to eventually run inferences locally, the network with the best results involves only three layers using the hyperbolic tangent activation function and the Adam Stochastic Gradient Descent (SGD) optimizer. The model is trained with a randomly-selected dataset using 640 vectors (80\% of the data) and tested with 160 vectors (20\%). Using the mean squared error as a loss function between the data and the model's prediction, we achieve a loss of only 1.47x10$^{-4}$ on the training set and 1.22x10$^{-4}$ on the test set. As such, we believe this appealing approach offers a new pathway towards significantly better datasets using readily-available ultra low-cost sensors. 
\end{abstract}

\begin{document}
\flushbottom
\maketitle

\thispagestyle{empty}

\section*{Introduction}
Artificial intelligence (AI) has been gradually changing medicine throughout the last few years. As exhibited by Bohr \& Memarzadeh \cite{bohr_chapter_nodate}, it can impact all areas of healthcare by enabling more precise disease detection, image analysis, patient monitoring, more efficient self-administration medication among others\cite{fouad_analyzing_2020}. Combining AI with health monitoring devices can significantly decrease healthcare costs\cite{bohr_chapter_nodate}. In particular, deep learning (DL) can help find hidden correlations and patterns using advanced machine-learning algorithms, including artificial neural networks (ANN)\cite{rawat_optimized_2022,jha_artificial_2022}. Thereby, deep machine learning can be used to achieve earlier recognition of complex patterns in a patient's data in order to detect anomalies and correlate symptoms \& diseases. This new branch of medicine can make it more accessible and affordable \cite{uwaoma_building_2021,mohd_ariff_establish_2019}. Exploiting machine learning, we seek to demonstrate that combinations of lower-cost \& lower-precision sensors can potentially become as precise as any cutting-edge healthcare technology, reducing costs and providing a more universal access to healthcare\cite{bohr_chapter_nodate}. Building on this philosophy, this report establishes how a deep learning approach can yield more accurate data predictions from an ultra low-cost temperature sensors array.

\subsection*{Temperature Sensors}
\noindent Temperature sensors come in many designs and materials, according to their target application. Beyond cost, the key features to consider are reliability, response time, accuracy, sensitivity, temperature range and, for skin temperature, wearability.\cite{childs_review_2000} They can include thermocouples, resistance temperature detectors (RTDs), thermistors and semiconductor sensors, each with their own advantages and disadvantages. \cite{khan_wearable_2021,rai_temperature_2007,tranca_precision_2018}. Detailed specifications for these sensors are found in the literature \cite{childs_review_2000,rai_temperature_2007}. For this project, we use negative temperature coefficient (NTC) thermistors and semiconductor-based integrated circuits (ICs). The NTC sensors measure change in resistance. The temperature is then computed with the Steinhart-Hart equation given as\cite{noauthor_temperature_nodate},

\begin{equation}\hspace{6cm}\frac{1}{T} = A + B\ln{R} + C(\ln{R})^{3}\end{equation}

\noindent where T is temperature in Kelvin, R is the thermistor resistance and A,B,C are constants specific to the sensor device, usually provided by manufacturers\cite{noauthor_temperature_nodate}. These are analog sensors where the output is a continuous electric signal that is converted to temperature. In contrast, integrated circuits (IC)-based sensors use bipolar transistors to do the measurements. The specific IC sensor chosen for this work also includes an analog-to-digital converter. As such, the output signal from the sensor is a non-continuous temperature reading.\\

\subsection*{Temperature in Medicine}
\noindent Body temperature is one of the key vital signs for health assessment. The normal temperature may slightly vary between individuals, but is considered normal when at 37$\degree$C\cite{ogoina_fever_2011}. However, the temperature will be different depending on the part of the body where it is measured, the ambient temperature and the subject's activities. Extremities tend to be colder\cite{ganong_regulation_2012}. The human body has built-in temperature regulation mechanisms in order to maintain its temperature at 37$\degree$C. These mechanisms can be cold-activated like shivering, hunger and goosebumps, or they can be heat-activated like sweating and accelerated breathing\cite{ganong_regulation_2012}. As such, temperature measurements are especially useful to detect infections and inflammations, but more generally for immune response detection. Bacterial discharge and virus load cause fever and can also be detected through a temperature increase\cite{huan_wearable_2022,ogoina_fever_2011}. Using wearable devices, it is now possible to continuously monitor several health indicators and vital signs outside the clinics\cite{dunn_wearable_2021,huan_wearable_2022}. For example, temperature imaging can be used to verify blood flow in order to detect small temperature changes on patients with vascular disorders\cite{philip_infrared_2009}. Another application is to monitor disease and treatment evolution through temperature, for example in pneumonia patients\cite{qu_low-cost_2022} or high-risk diabetic patients afflicted with foot ulcerations\cite{armstrong_skin_2007}. Furthermore, temperature can be used to distinguish between superficial and deep skin burns, but also to monitor and predict the skin regeneration and healing processes\cite{ganon_contribution_2020}. It is also used for the monitoring of infected wounds\cite{pusta_wearable_2021,zhang_flexible_2021}. Currently, thermal imaging is almost exclusively used in clinical settings. Although it offers many advantages as a simple non-invasive approach, its main disadvantage is that it works better directly on the skin\cite{qu_low-cost_2022}. This can cause accessibility, comfort and privacy issues, depending on the affected area. Furthermore, it can only be done in a medical environment. In contrast, temperature sensors can be easily implemented in wearable devices such as watches\cite{huan_wearable_2022}, patches\cite{nakata_wearable_2017} and even face masks\cite{laurino_innovative_2022} to be even less intrusive and allow to monitor the patients continuously. Although this can bring calibration and accuracy issues, we intend for our DL model to improve wearable devices and learn to make-up for these intrinsic issues.

\subsection*{Machine Learning}
\noindent The lower accuracies and precisions of low-cost printed sensors usually stem from less reliable designs, coupled with cheaper materials and fabrication techniques. As such, an interesting idea would be to exploit machine learning algorithms on a statistically-significant number of those low-cost sensors (array) in order to potentially learn to compensate for those design and fabrication flaws.

\noindent So far, only a handful of research groups have reported significant progress in this exciting new approach to further enhance the performances of inkjet printed sensors beyond their physical limits. In 2019, an array of 20 printed sensors coupled with a two-stage machine learning approach produced an artificial nose used for food classification\cite{schroeder_chemiresistive_2019}. There, a featurized-random forest or k-nearest neighbor classification (with similar accuracy) is first used for classification into food categories (ex: cheese, liquor, oil). Then, a combinatorial selector scan is used to class the specific food item within its category (ex: rum, vodka, whiskey, gin, and tequila as liquors) \cite{schroeder_chemiresistive_2019}. In 2020, researchers achieved sign-to-speech translation using machine-learning-assisted stretchable sensor arrays\cite{zhou_sign--speech_2020}. Meanwhile, biomolecular \& protein sensing \cite{pandit_machine_2019,behera_machine_2021}, as well as gases \& pollutant mixtures detection \cite{khan_nanowire-based_2020,thorson_using_2019} were also demonstrated using low-cost printed sensors with machine learning treatment. In 2020, a multi-disciplinary team developed a smartphone-based DNA diagnostic tool for malaria used in rural Uganda in order to also improve connectivity between such communities and centralized medical facilities \cite{guo_smartphone-based_2021}. They used low-cost paper-based microfluidic diagnostic test and had a disease detection accuracy over 98\% \cite{guo_smartphone-based_2021}. In all those cases, the idea was ultimately to equip these low-cost sensor arrays with some intelligence in order to perform a certain \textit{classification} task\cite{schroeder_chemiresistive_2019,zhou_sign--speech_2020,pandit_machine_2019,behera_machine_2021,khan_nanowire-based_2020,thorson_using_2019,tsakanikas_machine_2020}.

\noindent This work demonstrates that low-cost and low-precision temperature sensors can be used in combination with deep machine learning frameworks to yield more precise and more accurate temperature predictions. This opens a wide range of applications in the medical field, where sensors could be placed in different body parts, but maybe still predict the  body temperature.

\section*{Materials and Methods}

\subsection*{Data Collection}
\subsubsection*{Materials}
In order to create our sensor array, we chose two types of low-cost temperature sensors. We used 16 digital temperature sensors and 16 analog temperature sensors, all with accuracies between 0.5 to 2.0$^{\circ}$C.

\noindent To collect the data from our sensors, we use an Arduino Mega2650\cite{arduino_datasheet_nodate} and Arduino's IDE. The Mega2650 offers 54 digital inputs and 16 analog inputs, which is sufficient to connect all our sensors to the microcontroller. It operates at 5V and 16 MHz frequency.
\subsubsection*{Methods}
The sensors are mounted on an IKA C-MAG HS 7 control hotplate \cite{ika_datasheet_nodate} with thermal paste to ensure conductivity and fixed with thermally-conductive tape, using the configurations shown in Figures 1 \& 2. The sensors are placed in a 4x8 array, covering the center of the plate as indicated in Figure 1(a). Two (2) rows consist of the digital sensors and two (2) rows for the  analog  sensors. This specific configuraton is only chosen to facilitate wire-management and data-collection. 

\noindent Using two types of sensors helps diversify our dataset, while providing enough data for a meaningful distribution (see Supplementary Information section to compare both sensor types).    All 32 sensors are connected through wires to the Arduino microcontroller as shown in  Figure 1. There, Figure 1(a) shows a representation of the experimental setup condition. The approximate placement of the 32 low-cost temperature sensors is schematically represented atop a thermal image of the hotplate when set at 50$\degree$C. From the thermal image acquired using a FLIR-One infrared camera, it is clear that the temperature is not uniform everywhere on the hotplate. This configuration was chosen on purpose to represent the human body, where the temperature can fluctuate quite a bit from one location to another. We cycled the hotplate's set-temperature from 30 to 45$\degree$C, with increments of 1$\degree$C as a staircase signal. The heating plate's accuracy is $\pm 0.15\degree$C according to specifications. As the temperature varies quite significantly over the surface, this parameter isn't really meaningful in this study. Instead, we want our model to learn to predict the hotplate's set temperature with the highest precision. Similarly, the body temperature can also vary significantly from one point to another. When measured orally, this temperature is established as true, whereas elsewhere it may be different. Thus, we want our model learn to predict the established temperature of the plate, even if it is changing wildly depending on the position of the sensors on the hotplate. 

\noindent Figure 1(b) shows the circuit on the Arduino Mega2650. For each degree measurement, we waited for two minutes to make sure that the hotplate's temperature is stable before collecting the sensors' readings each 1.5 seconds for four minutes using the Arduino. Afterwards, 50 random vectors were chosen from the entire sample. Thus, our dataset consists of 50 vectors of 32 components for each temperature degree, for a total of 800 vectors. All vectors are then labeled according to the plate's set temperature and accuracy.

\begin{figure}[h]
     \centering
     \begin{subfigure}[b]{0.45\textwidth}
         \centering
         \includegraphics[width=5cm]{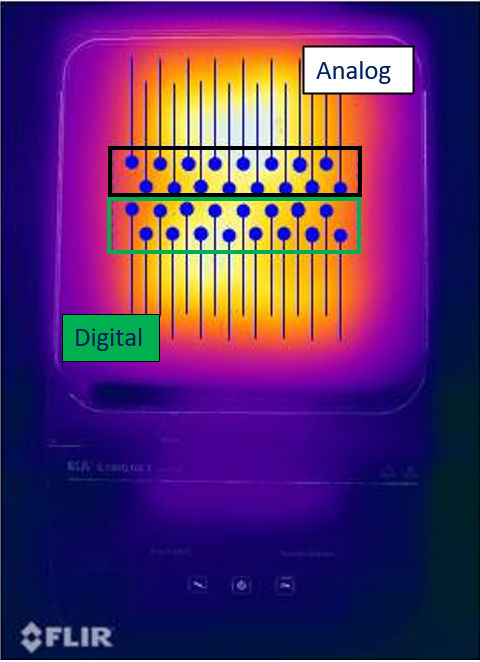}
         \caption{}

     \end{subfigure}
     \hfill
     \begin{subfigure}[b]{0.45\textwidth}
         \centering
         \includegraphics[width=\textwidth]{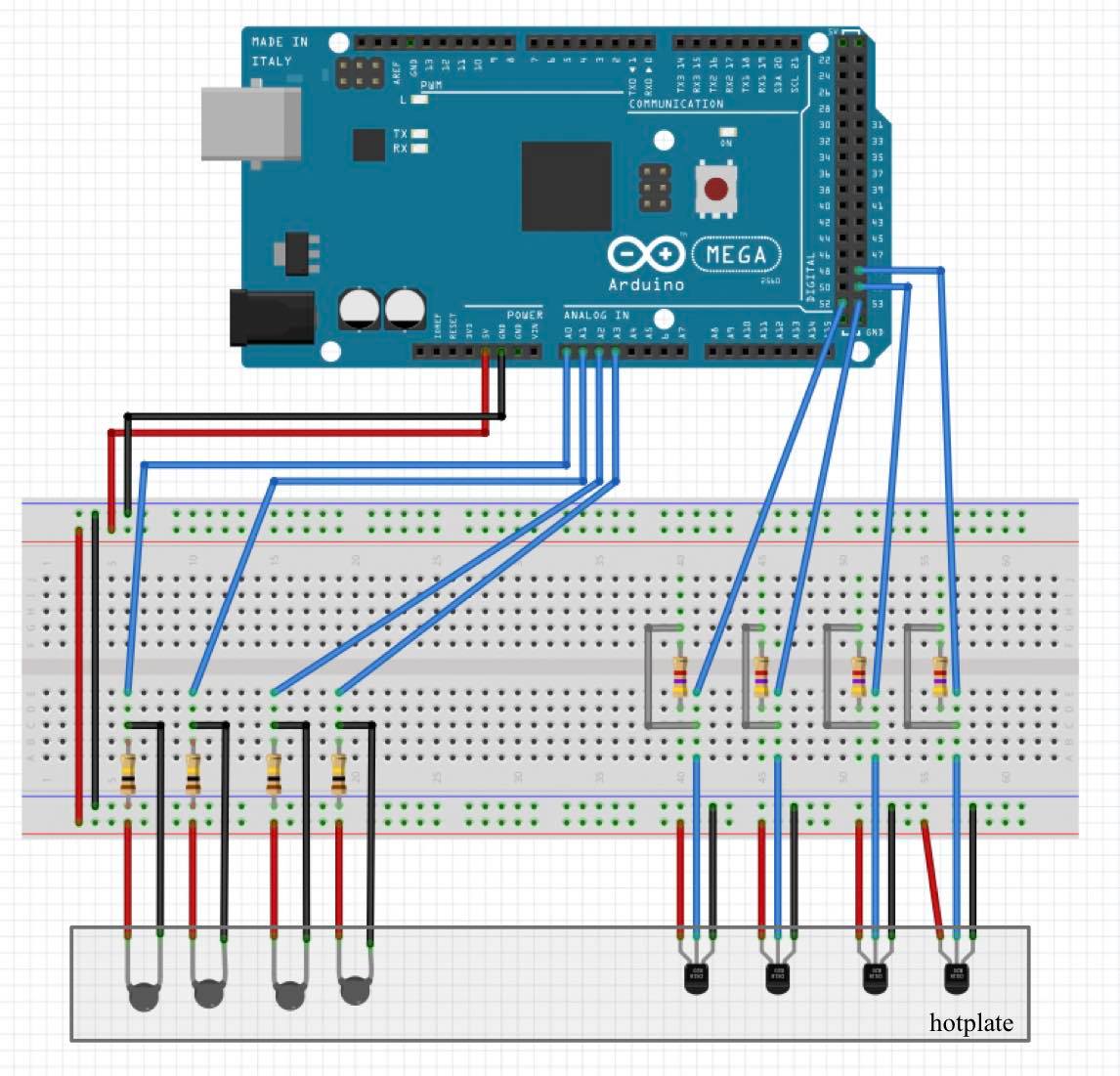}
         \caption{}

     \end{subfigure}
        \caption{(a) Placement of the 32 low-cost temperature sensors schematically represented atop a thermal image of the hotplate when set at 50$\degree$C. The two rows bounded in the black box are analog sensors, where as the green box contains digital sensors. (b) Schematic of our circuit. Only four sensors of each type are shown to simplify the figure.}
        \label{fig:three graphs}
\end{figure}

\subsection*{Neural Network} 
\subsubsection*{Methods}

A feed-forward artificial neural network is a deep learning algorithm generally using large quantities of data to learn to recognize hidden patterns in order to make more accurate predictions\cite{li_application_2022}. In this case, we designed a regression model, which predicts an output based off the data you feed it. The architecture of an artificial neural network is composed of interconnected neuron layers. It also contains an input layer, which forwards the data to the hidden layers for computational purpose. The information is then spread to the output layer, making a final prediction. 
Here, we used a supervised training method.\cite{pattanayak_pro_nodate} To do so, the data set is randomly split in two subsets. 80\% went to the training set, while the remaining 20\% composed the validation set. To design our model, we use TensorFlow's module Keras\cite{chollet_deep_2021}. We chose a three-layer deep neural network (DNN) shown in Figure 3. Its hidden layer has 20 neurons and all are activated with an hyperbolic tangent function $(\tanh(x)=\frac{e^{x} - e^{-x}}{e^{x} + e^{-x}})$. We chose the Adam optimizer with a 0.01 learning rate. The computed loss function for our regression is the mean squared error $(MSE =\frac{1}{n}  \sum_{j=1}^n (Y_{j}-\hat{Y}_{j})^2 )$. We obtained a minimal loss by training the data set with 300 epochs. (See Supplementary Information for optimization details on the algorithm)

\begin{figure}[h]

 \centering
    \includegraphics[width=0.60\textwidth]{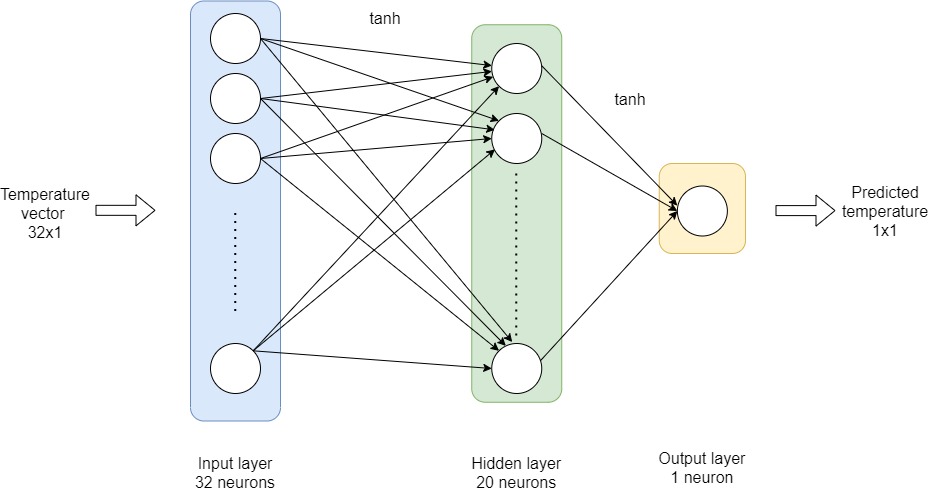}
    \caption{Optimized deep neural network (DNN) architecture used to learn to predict the (set) temperature from the read-out data of our 4 x 8 array of low-cost and low-accuracy temperature sensors.}
    \label{fig:my_label}
\end{figure}

\noindent In this project, the model trains itself with 640 temperature vectors composed from the readings of the 32 sensors and labelled with the set temperature of the hotplate. Using this training set, the model learns to predict the set temperature of the hotplate (labels) from the patterns and features of the array. Once the model is trained, we evaluate its performances using the test vectors and evaluate its set-temperature predictions from the sensors' data. The predictions according to the set temperature and compared with the sensors' readings are shown in Figure 4.

\section*{Results and Discussion}

\noindent As seen in Figure 3, the recorded temperatures fluctuate wildly from one sensor to another and the averaged sensor readings tend to increasingly underestimate the actual (set) temperature as the temperature is increased. For a set temperature of 45$\degree$C, some sensors read temperatures as low as 33$\degree$C. After only a few iterations of training, we observe that the DNN model architecture shown in Figure 3 can predict the actual (set) temperature with a 0.12$\degree$C accuracy using extremely low-quality sensing devices. These results suggest that the DNN learns to compensate extremely well for the poor sensors' precision and accuracy, as well as for the non-homogeneous temperature profile of the plate.
\\

    \begin{figure*}[h!]
        \centering
      
        \includegraphics[width=0.65\textwidth]{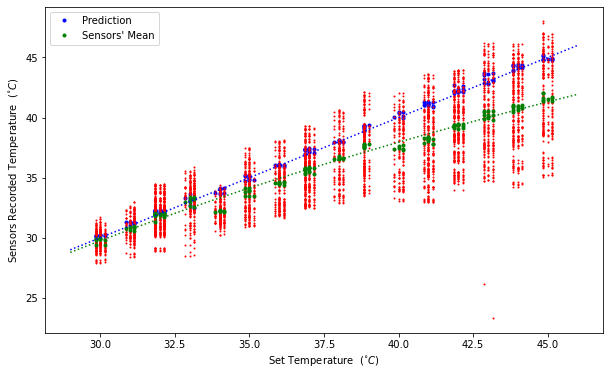}

        \caption{Individual temperature readings from each of the 32 sensors (\textcolor{red}{\textbullet}), mean value of all readings (\textcolor{teal}{\textbullet}) and the model's predictions (\textcolor{blue}{\textbullet}) as a function of the hotplate's set temperature. The dashed blue line indicates the target (set) temperature value.}
        \label{fig2}

    \end{figure*}

\begin{figure}[h!]
\begin{minipage}{0.45\textwidth}
\includegraphics[height=6cm]{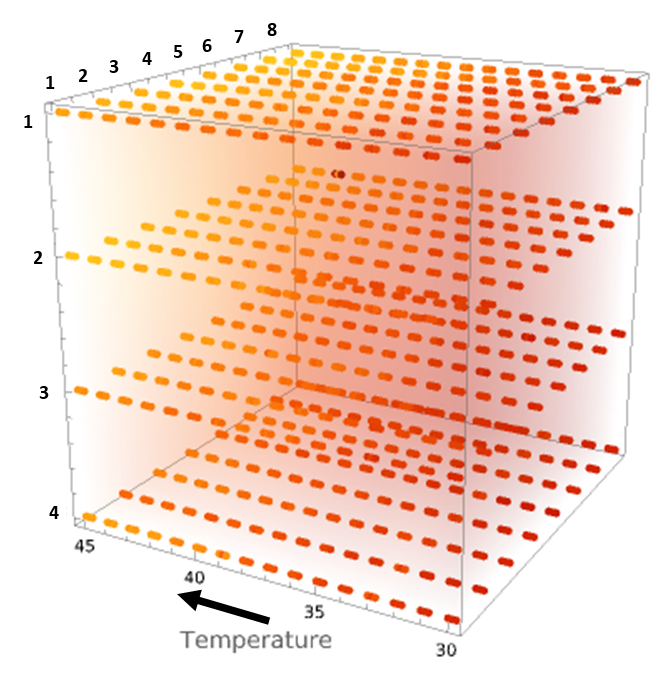}
\end{minipage}
\begin{minipage}{0.1\textwidth}
\includegraphics[height=5cm]{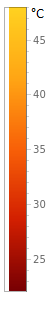}
\end{minipage}
\begin{minipage}{0.45\textwidth}
\includegraphics[height=6cm]{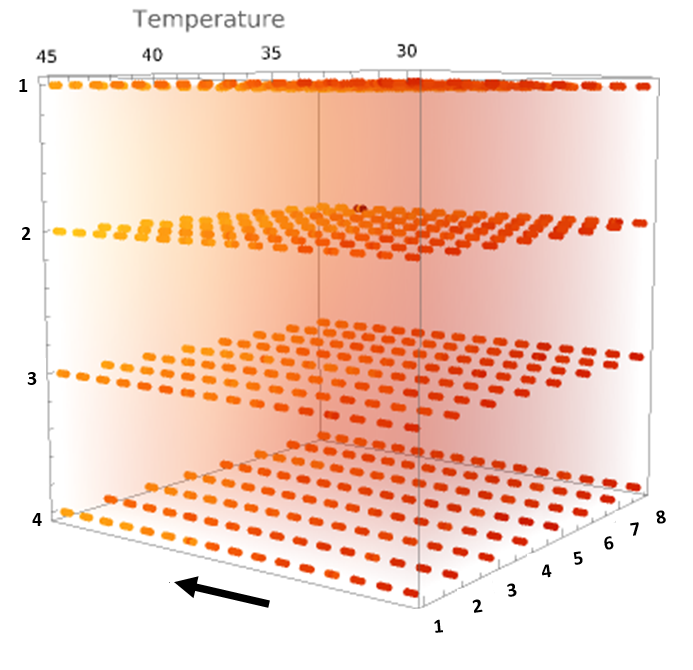}
\end{minipage}

\caption{3D heatmaps of our individual sensors' readings with increasing temperature in the black arrow's direction from two (2) different viewpoints. Each dotted line represents the evolution of one sensor. Axes going from 1 to 4 (row) and 1-8 (column) indicate the sensors' individual placement.}
\label{fig}
\end{figure}

\noindent Results from Figure 3 clearly highlight the sensors' large margin for error. The red vertical dotted lines indicate each sensors' readings according to each set temperature. As the set temperature increases, it is clear that lower temperature readings tend to move further from the actual set temperature than the maximal one. As such, the mean value extracted from the 32 sensors increasingly underestimates the set temperature. This increasing margin of error as the temperature rises can be explained by (1) the sensors' low-precision and (2) the non-uniformity of the hotplate's thermal profile for higher temperatures. The blue data points show the model's prediction, which accurately follows the set temperature. The sensors' readings' mean rather follows a logarithmic regression.
\\

\noindent Even if the hotplate's thermal profile becomes non-uniform as the temperature rises, our algorithm still performs very well. This is because the model also learns the behavior of each sensor according to its placement. Figure 4 shows a 3D representation of our 4x8 sensor array as the temperature increases on the hotplate (heatmap) from two (2) different viewpoints. Just like the thermal image of Figure 1, it is clear that some sensors are subjected to different temperatures locally. At 30$\degree$, the whole array is red (lower temperatures), and it then becomes increasingly yellow (higher temperatures) at different rates. It is clear from these results that the temperature is not uniform everywhere on the hotplate. Indeed, this experimental system was selected on purpose in order to accurately mimic the non-uniform temperatures experienced across the human body. 

\noindent \begin{minipage}{0.6\textwidth}
As mentioned earlier, the body temperature can vary significantly depending on where the measurement is taken, just like the sensors in our work. However, our model still manages to predict accurately the hotplate's reference (set) temperature, which would be the core body temperature in a clinical setting. Thus, we could place sensors over a patient's body and be able to know the core body temperature. This could also be helpful in creating a heat map of the patient's body. For instance, becomes possible to accurately predict the temperature at point A from a reading at point B, as shown in Figure 5, where the algorithm could predict the oral temperature from the wrist sensor.  But most importantly, it helps to get more accurate temperature readings from low-cost sensors.  Even though these timely results present an appealing proof-of-concept, this study also represents a strong foundation for future investigations. For example, one could try different types of sensors an/or modify the positioning and shape of the array. We already did a first try of data augmentation and randomized the test set to get an idea of the effect's on the prediction, as discussed in the \textit{Ablation Study} below. The chosen temperature range for this project is convenient for skin temperature measurements, but we could also try to expand it and accommodate the model's behavior in order to target other applications.
\vfill
\end{minipage}
\begin{minipage}{0.4\textwidth}
\centering
\includegraphics[width=0.60\textwidth]{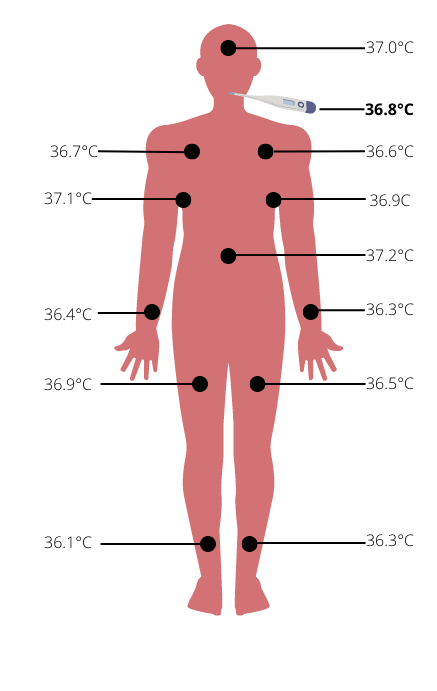}
\centering
\captionof{figure}{Representation of our model's transition onto the human body}
\label{fig:sample_figure}
\end{minipage}
\subsection*{Ablation Study}
In order to develop the best model for this project, we tried different parameters for the algorithm. This section describes the model's optimization process.

\subsubsection*{Loss and Activation functions }
 We compared the results using different loss functions for our regression model. We tried the mean absolute error (Figure 6a) and the mean squared logarithmic error and the root mean squared error (Figure 6b) resulting in loss values of 2.79x10$^{-2}$ and 4.57x10$^{-2}$ respectively. As shown in Figure 6(c), the mean squared error yields significantly better results. Furthermore, the combination of the hyperbolic tangent activation function for both layers results in the best precision.

\begin{figure}[h]
\begin{subfigure}{.5\textwidth}
  \centering
  \includegraphics[width=.9\linewidth]{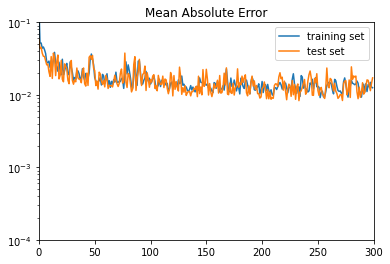}  
  \caption{}
  \label{fig:sub-first}
\end{subfigure}
\begin{subfigure}{.5\textwidth}
  \centering
  \includegraphics[width=.9\linewidth]{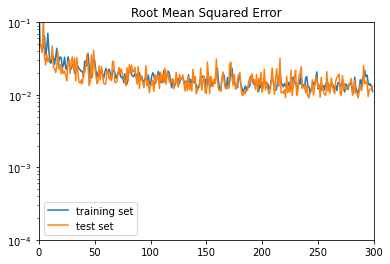}  
  \caption{}
  \label{fig:sub-second}
\end{subfigure}\\[1ex]
\begin{subfigure}{\linewidth}
\centering
\includegraphics[width=.45\linewidth]{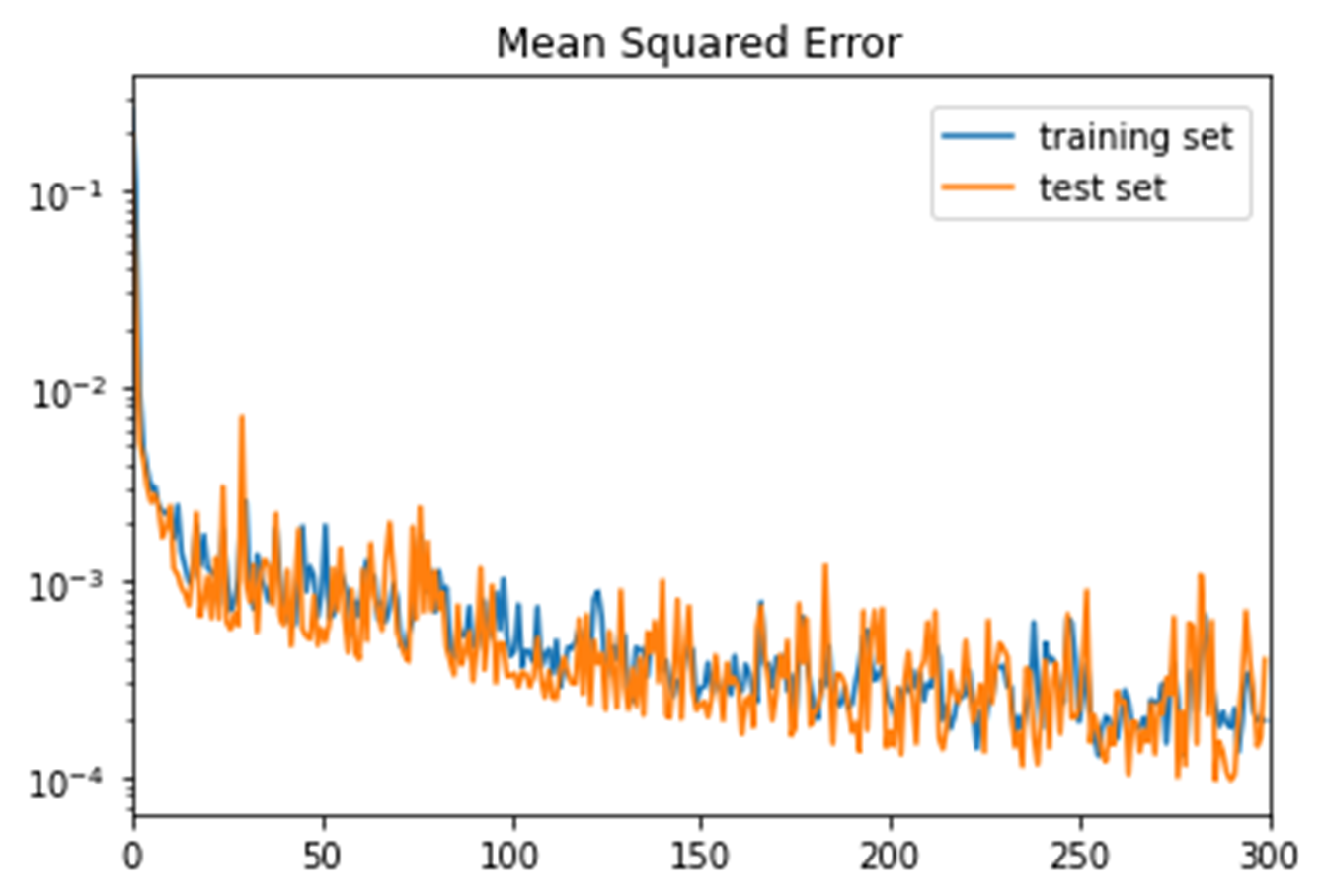}
\caption{}
\label{fig:sub3}
\end{subfigure}
\caption{Results for different loss functions used to optimize our model in a logarithmic scale. (a) mean absolute error (MAE), (b) root mean squared error (RMSE) and (c) mean squared error (MSE), all with hyperbolic tangent activation function.}
\label{fig:test}
\end{figure}

\begin{figure}[h!]
\begin{subfigure}{.5\textwidth}
  \centering
  \includegraphics[width=.9\linewidth]{figures/logscale.png}  
  \caption{}
  \label{fig:sub-first}
\end{subfigure}
\begin{subfigure}{.5\textwidth}
  \centering
  \includegraphics[width=.9\linewidth]{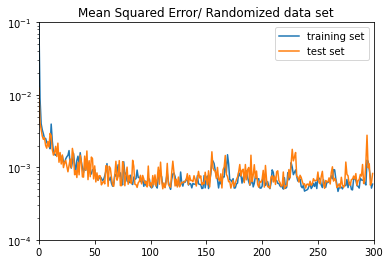}  
  \caption{}
  \label{fig:sub-second}
\end{subfigure}
\caption{(a) MSE loss function on logarithmic scale for our model (b) MSE loss function on logarithmic scale for a model trained using a completely randomized data set}
\label{fig:test}
\end{figure}

\subsubsection*{Randomized vectors}
If we shuffle the components of the testing vectors, the algorithm doesn't perform as well. This is because part of the algorithm is also learning the heat profile of the hotplate, as it is intended. For instance, a shuffled test set fed into our model achieves a MSE of 6.58x10$^{-3}$, which is roughly a factor thirty (30$\times$) larger than our best MSE result. This is due to the fact that when shuffling the components of each vector, we mix the heat map of our dataset. Therefore, the model can't rely on that specific property to predict the temperature, and that's why it becomes a bit less precise.
\\
As a result, we also tried to train our model using randomly-shuffled vectors. Each one was randomly shuffled so that the mapping aspect is completely eliminated. The same model computed a 6.97x10$^{-4}$ of the test set, which is only a factor five (5$\times$)  higher than our best MSE results. Comparison for both models is shown in Figure 7.

\section*{Conclusion}
While it is important to continue to improve materials and fabrication processes to improve sensors, the impact that artificial intelligence can also have should not be overlooked. This paper clearly demonstrates that machine learning algorithms are a powerful way to rapidly enhance readily-available low-cost sensor performances and to generate heatmaps from low-cost sensor arrays. Research is currently oriented towards implementing some intelligence onto sensor arrays in order to perform classification tasks. We show that applying such deep-learning algorithms to low-cost sensor arrays can also improve their performances. In the future, it will be interesting to try this approach to identify the best designs to measure body temperatures. To do so, researchers will need to consider the sensor's biocompatibility with skin\cite{patel_drawn--skin_2022}. This system has a great potential for medical applications for non-invasive temperature measurements. As it is currently done with biomarkers\cite{pusta_wearable_2021}, such sensor arrays could also be used to monitor inflammatory responses or recovery in burned and wounded patients. The advantage our sensors is that they require no chemical reactions and aren't disposable. Furthermore, the thermal map generated by our model could be helpful to monitor the evolution of wounds and injuries, at much lower costs than current methods. 

\section*{Data availability}
The datasets generated during and/or analysed during the current study are available from the corresponding author on reasonable request.
\newpage
\bibliography{references}

\begin{thebibliography}{10}
\urlstyle{rm}
\expandafter\ifx\csname url\endcsname\relax
  \def\url#1{\texttt{#1}}\fi
\expandafter\ifx\csname urlprefix\endcsname\relax\def\urlprefix{URL }\fi
\expandafter\ifx\csname doiprefix\endcsname\relax\def\doiprefix{DOI: }\fi
\providecommand{\bibinfo}[2]{#2}
\providecommand{\eprint}[2][]{\url{#2}}

\bibitem{bohr_chapter_nodate}
\bibinfo{author}{Bohr, A.} \& \bibinfo{author}{Memarzadeh, K.}
\newblock \bibinfo{title}{Chapter 2. {The} rise of artificial intelligence in
  healthcare applications {\textbar} {Elsevier} {Enhanced} {Reader}},
  \doiprefix\url{10.1016/B978-0-12-818438-7.00002-2}.

\bibitem{fouad_analyzing_2020}
\bibinfo{author}{Fouad, H.}, \bibinfo{author}{Hassanein, A.~S.},
  \bibinfo{author}{Soliman, A.~M.} \& \bibinfo{author}{Al-Feel, H.}
\newblock \bibinfo{journal}{\bibinfo{title}{Analyzing patient health
  information based on {IoT} sensor with {AI} for improving patient assistance
  in the future direction}}.
\newblock {\emph{\JournalTitle{Measurement}}} \textbf{\bibinfo{volume}{159}},
  \bibinfo{pages}{107757}, \doiprefix\url{10.1016/j.measurement.2020.107757}
  (\bibinfo{year}{2020}).

\bibitem{rawat_optimized_2022}
\bibinfo{author}{Rawat, D.}, \bibinfo{author}{{Meenakshi}},
  \bibinfo{author}{Pawar, L.}, \bibinfo{author}{Bathla, G.} \&
  \bibinfo{author}{Kant, R.}
\newblock \bibinfo{title}{Optimized {Deep} {Learning} {Model} for {Lung}
  {Cancer} {Prediction} {Using} {ANN} {Algorithm}}.
\newblock In \emph{\bibinfo{booktitle}{2022 3rd {International} {Conference} on
  {Electronics} and {Sustainable} {Communication} {Systems} ({ICESC})}},
  \bibinfo{pages}{889--894}, \doiprefix\url{10.1109/ICESC54411.2022.9885607}
  (\bibinfo{year}{2022}).

\bibitem{jha_artificial_2022}
\bibinfo{author}{Jha, K.~K.}, \bibinfo{author}{Das, P.} \&
  \bibinfo{author}{Dutta, H.~S.}
\newblock \bibinfo{title}{Artificial {Neural} {Network}-{Based} {Leukaemia}
  {Identification} and {Prediction} using {Ensemble} {Deep} {Learning}
  {Model}}.
\newblock In \emph{\bibinfo{booktitle}{2022 {International} {Conference} on
  {Communication}, {Computing} and {Internet} of {Things} ({IC3IoT})}},
  \bibinfo{pages}{1--6}, \doiprefix\url{10.1109/IC3IOT53935.2022.9767874}
  (\bibinfo{year}{2022}).

\bibitem{uwaoma_building_2021}
\bibinfo{author}{Uwaoma, C.} \& \bibinfo{author}{Mansingh, G.}
\newblock \bibinfo{title}{Building a {Decision} {Support} {System} for
  {Automated} {Mobile} {Asthma} {Monitoring} in {Remote} {Areas}},
  \doiprefix\url{10.48550/arXiv.2112.11195} (\bibinfo{year}{2021}).
\newblock \bibinfo{note}{ArXiv:2112.11195 [cs] version: 1}.

\bibitem{mohd_ariff_establish_2019}
\bibinfo{author}{Mohd~Ariff, M. A.~I.}, \bibinfo{author}{Then, Y.~L.} \&
  \bibinfo{author}{Tay, F.~S.}
\newblock \bibinfo{title}{Establish {Connection} {Between} {Remote} {Areas} and
  {City} to {Improve} {Healthcare} {Services}}.
\newblock In \emph{\bibinfo{booktitle}{2019 {International} {Conference} on
  {Green} and {Human} {Information} {Technology} ({ICGHIT})}},
  \bibinfo{pages}{18--23}, \doiprefix\url{10.1109/ICGHIT.2019.00012}
  (\bibinfo{year}{2019}).

\bibitem{childs_review_2000}
\bibinfo{author}{Childs, P.}, \bibinfo{author}{Greenwood, J.} \&
  \bibinfo{author}{Long, C.}
\newblock \bibinfo{journal}{\bibinfo{title}{Review of temperature
  measurement}}.
\newblock {\emph{\JournalTitle{Review of Scientific Instruments}}}
  \textbf{\bibinfo{volume}{71}}, \bibinfo{pages}{2959--78},
  \doiprefix\url{10.1063/1.1305516} (\bibinfo{year}{2000}).
\newblock \bibinfo{note}{Place: USA Publisher: AIP}.

\bibitem{khan_wearable_2021}
\bibinfo{author}{Khan, S.}, \bibinfo{author}{Ali, S.}, \bibinfo{author}{Khan,
  A.} \& \bibinfo{author}{Bermak, A.}
\newblock \bibinfo{journal}{\bibinfo{title}{Wearable {Printed} {Temperature}
  {Sensors}: {Short} {Review} on {Latest} {Advances} for {Biomedical}
  {Applications}}}.
\newblock {\emph{\JournalTitle{IEEE Reviews in Biomedical Engineering}}}
  \bibinfo{pages}{1--1}, \doiprefix\url{10.1109/RBME.2021.3121480}
  (\bibinfo{year}{2021}).
\newblock \bibinfo{note}{Conference Name: IEEE Reviews in Biomedical
  Engineering}.

\bibitem{rai_temperature_2007}
\bibinfo{author}{Rai, V.~K.}
\newblock \bibinfo{journal}{\bibinfo{title}{Temperature sensors and optical
  sensors}}.
\newblock {\emph{\JournalTitle{Applied Physics B}}}
  \textbf{\bibinfo{volume}{88}}, \bibinfo{pages}{297--303},
  \doiprefix\url{10.1007/s00340-007-2717-4} (\bibinfo{year}{2007}).

\bibitem{tranca_precision_2018}
\bibinfo{author}{Trancã, D.-C.} \emph{et~al.}
\newblock \bibinfo{title}{Precision and linearity of analog temperature sensors
  for industrial {IoT} devices}.
\newblock In \emph{\bibinfo{booktitle}{2018 17th {RoEduNet} {Conference}:
  {Networking} in {Education} and {Research} ({RoEduNet})}},
  \bibinfo{pages}{1--6}, \doiprefix\url{10.1109/ROEDUNET.2018.8514122}
  (\bibinfo{year}{2018}).
\newblock \bibinfo{note}{ISSN: 2247-5443}.

\bibitem{noauthor_temperature_nodate}
\bibinfo{title}{Temperature {Sensor} {Tutorial} - {Maxim} {\textbar}
  {DigiKey}}.

\bibitem{ogoina_fever_2011}
\bibinfo{author}{Ogoina, D.}
\newblock \bibinfo{journal}{\bibinfo{title}{Fever, fever patterns and diseases
  called ‘fever’ – {A} review}}.
\newblock {\emph{\JournalTitle{Journal of Infection and Public Health}}}
  \textbf{\bibinfo{volume}{4}}, \bibinfo{pages}{108--124},
  \doiprefix\url{10.1016/j.jiph.2011.05.002} (\bibinfo{year}{2011}).

\bibitem{ganong_regulation_2012}
\bibinfo{author}{Ganong, W.~F.}, \bibinfo{author}{Barrett, K.~E.},
  \bibinfo{author}{Barman, S.~M.}, \bibinfo{author}{Boitano, S.} \&
  \bibinfo{author}{Brooks, H.~L.}
\newblock \bibinfo{title}{La régulation hypothalamique des fonctions
  hormonales}.
\newblock In \emph{\bibinfo{booktitle}{Physiologie médicale}},
  \bibinfo{pages}{273--288} (\bibinfo{publisher}{De Boeck},
  \bibinfo{year}{2012}), \bibinfo{edition}{3e} edn.

\bibitem{huan_wearable_2022}
\bibinfo{author}{Huan, J.} \emph{et~al.}
\newblock \bibinfo{journal}{\bibinfo{title}{A {Wearable} {Skin} {Temperature}
  {Monitoring} {System} for {Early} {Detection} of {Infections}}}.
\newblock {\emph{\JournalTitle{IEEE Sensors Journal}}}
  \textbf{\bibinfo{volume}{22}}, \bibinfo{pages}{1670--1679},
  \doiprefix\url{10.1109/JSEN.2021.3131500} (\bibinfo{year}{2022}).
\newblock \bibinfo{note}{Conference Name: IEEE Sensors Journal}.

\bibitem{dunn_wearable_2021}
\bibinfo{author}{Dunn, J.} \emph{et~al.}
\newblock \bibinfo{journal}{\bibinfo{title}{Wearable sensors enable
  personalized predictions of clinical laboratory measurements}}.
\newblock {\emph{\JournalTitle{Nature Medicine}}}
  \textbf{\bibinfo{volume}{27}}, \bibinfo{pages}{1105--1112},
  \doiprefix\url{10.1038/s41591-021-01339-0} (\bibinfo{year}{2021}).
\newblock \bibinfo{note}{Number: 6 Publisher: Nature Publishing Group}.

\bibitem{philip_infrared_2009}
\bibinfo{author}{Philip, J.} \emph{et~al.}
\newblock \bibinfo{journal}{\bibinfo{title}{Infrared thermal imaging for
  detection of peripheral vascular disorders}}.
\newblock {\emph{\JournalTitle{Journal of Medical Physics}}}
  \textbf{\bibinfo{volume}{34}}, \bibinfo{pages}{43},
  \doiprefix\url{10.4103/0971-6203.48720} (\bibinfo{year}{2009}).

\bibitem{qu_low-cost_2022}
\bibinfo{author}{Qu, Y.}, \bibinfo{author}{Meng, Y.}, \bibinfo{author}{Fan, H.}
  \& \bibinfo{author}{Xu, R.~X.}
\newblock \bibinfo{journal}{\bibinfo{title}{Low-cost thermal imaging with
  machine learning for non-invasive diagnosis and therapeutic monitoring of
  pneumonia}}.
\newblock {\emph{\JournalTitle{Infrared Physics \& Technology}}}
  \textbf{\bibinfo{volume}{123}}, \bibinfo{pages}{104201},
  \doiprefix\url{10.1016/j.infrared.2022.104201} (\bibinfo{year}{2022}).

\bibitem{armstrong_skin_2007}
\bibinfo{author}{Armstrong, D.~G.} \emph{et~al.}
\newblock \bibinfo{journal}{\bibinfo{title}{Skin {Temperature} {Monitoring}
  {Reduces} the {Risk} for {Diabetic} {Foot} {Ulceration} in {High}-risk
  {Patients}}}.
\newblock {\emph{\JournalTitle{The American Journal of Medicine}}}
  \textbf{\bibinfo{volume}{120}}, \bibinfo{pages}{1042--1046},
  \doiprefix\url{10.1016/j.amjmed.2007.06.028} (\bibinfo{year}{2007}).

\bibitem{ganon_contribution_2020}
\bibinfo{author}{Ganon, S.}, \bibinfo{author}{Guédon, A.},
  \bibinfo{author}{Cassier, S.} \& \bibinfo{author}{Atlan, M.}
\newblock \bibinfo{journal}{\bibinfo{title}{Contribution of thermal imaging in
  determining the depth of pediatric acute burns}}.
\newblock {\emph{\JournalTitle{Burns}}} \textbf{\bibinfo{volume}{46}},
  \bibinfo{pages}{1091--1099}, \doiprefix\url{10.1016/j.burns.2019.11.019}
  (\bibinfo{year}{2020}).

\bibitem{pusta_wearable_2021}
\bibinfo{author}{Pusta, A.}, \bibinfo{author}{Tertiș, M.},
  \bibinfo{author}{Cristea, C.} \& \bibinfo{author}{Mirel, S.}
\newblock \bibinfo{journal}{\bibinfo{title}{Wearable {Sensors} for the
  {Detection} of {Biomarkers} for {Wound} {Infection}}}.
\newblock {\emph{\JournalTitle{Biosensors}}} \textbf{\bibinfo{volume}{12}},
  \bibinfo{pages}{1}, \doiprefix\url{10.3390/bios12010001}
  (\bibinfo{year}{2021}).

\bibitem{zhang_flexible_2021}
\bibinfo{author}{Zhang, Y.} \emph{et~al.}
\newblock \bibinfo{journal}{\bibinfo{title}{Flexible integrated sensing
  platform for monitoring wound temperature and predicting infection}}.
\newblock {\emph{\JournalTitle{Microbial Biotechnology}}}
  \textbf{\bibinfo{volume}{14}}, \bibinfo{pages}{1566--1579},
  \doiprefix\url{10.1111/1751-7915.13821} (\bibinfo{year}{2021}).

\bibitem{nakata_wearable_2017}
\bibinfo{author}{Nakata, S.}, \bibinfo{author}{Arie, T.},
  \bibinfo{author}{Akita, S.} \& \bibinfo{author}{Takei, K.}
\newblock \bibinfo{journal}{\bibinfo{title}{Wearable, {Flexible}, and
  {Multifunctional} {Healthcare} {Device} with an {ISFET} {Chemical} {Sensor}
  for {Simultaneous} {Sweat} {pH} and {Skin} {Temperature} {Monitoring}}}.
\newblock {\emph{\JournalTitle{ACS Sensors}}} \textbf{\bibinfo{volume}{2}},
  \bibinfo{pages}{443--448}, \doiprefix\url{10.1021/acssensors.7b00047}
  (\bibinfo{year}{2017}).
\newblock \bibinfo{note}{Publisher: American Chemical Society}.

\bibitem{laurino_innovative_2022}
\bibinfo{author}{Laurino, M.} \emph{et~al.}
\newblock \bibinfo{title}{An {Innovative} {Sensorized} {Face} {Mask} for
  {Early} {Detection} of {Physiological} {Changes} {Associated} with {Viral}
  {Infection}}.
\newblock In \emph{\bibinfo{booktitle}{2022 44th {Annual} {International}
  {Conference} of the {IEEE} {Engineering} in {Medicine} \& {Biology} {Society}
  ({EMBC})}}, \bibinfo{pages}{933--936},
  \doiprefix\url{10.1109/EMBC48229.2022.9871775} (\bibinfo{year}{2022}).
\newblock \bibinfo{note}{ISSN: 2694-0604}.

\bibitem{schroeder_chemiresistive_2019}
\bibinfo{author}{Schroeder, V.} \emph{et~al.}
\newblock \bibinfo{journal}{\bibinfo{title}{Chemiresistive {Sensor} {Array} and
  {Machine} {Learning} {Classification} of {Food}}}.
\newblock {\emph{\JournalTitle{ACS Sensors}}} \textbf{\bibinfo{volume}{4}},
  \bibinfo{pages}{2101--2108}, \doiprefix\url{10.1021/acssensors.9b00825}
  (\bibinfo{year}{2019}).
\newblock \bibinfo{note}{Publisher: American Chemical Society}.

\bibitem{zhou_sign--speech_2020}
\bibinfo{author}{Zhou, Z.} \emph{et~al.}
\newblock \bibinfo{journal}{\bibinfo{title}{Sign-to-speech translation using
  machine-learning-assisted stretchable sensor arrays}}.
\newblock {\emph{\JournalTitle{Nature Electronics}}}
  \textbf{\bibinfo{volume}{3}}, \bibinfo{pages}{571--578},
  \doiprefix\url{10.1038/s41928-020-0428-6} (\bibinfo{year}{2020}).
\newblock \bibinfo{note}{Number: 9 Publisher: Nature Publishing Group}.

\bibitem{pandit_machine_2019}
\bibinfo{author}{Pandit, S.}, \bibinfo{author}{Banerjee, T.},
  \bibinfo{author}{Srivastava, I.}, \bibinfo{author}{Nie, S.} \&
  \bibinfo{author}{Pan, D.}
\newblock \bibinfo{journal}{\bibinfo{title}{Machine {Learning}-{Assisted}
  {Array}-{Based} {Biomolecular} {Sensing} {Using} {Surface}-{Functionalized}
  {Carbon} {Dots}}}.
\newblock {\emph{\JournalTitle{ACS Sensors}}} \textbf{\bibinfo{volume}{4}},
  \bibinfo{pages}{2730--2737}, \doiprefix\url{10.1021/acssensors.9b01227}
  (\bibinfo{year}{2019}).
\newblock \bibinfo{note}{Publisher: American Chemical Society}.

\bibitem{behera_machine_2021}
\bibinfo{author}{Behera, P.} \emph{et~al.}
\newblock \bibinfo{journal}{\bibinfo{title}{Machine {Learning}-{Assisted}
  {Array}-{Based} {Detection} of {Proteins} in {Serum} {Using} {Functionalized}
  {MoS2} {Nanosheets} and {Green} {Fluorescent} {Protein} {Conjugates}}}.
\newblock {\emph{\JournalTitle{ACS Applied Nano Materials}}}
  \textbf{\bibinfo{volume}{4}}, \bibinfo{pages}{3843--3851},
  \doiprefix\url{10.1021/acsanm.1c00244} (\bibinfo{year}{2021}).
\newblock \bibinfo{note}{Publisher: American Chemical Society}.

\bibitem{khan_nanowire-based_2020}
\bibinfo{author}{Khan, M. A.~H.}, \bibinfo{author}{Thomson, B.},
  \bibinfo{author}{Debnath, R.}, \bibinfo{author}{Motayed, A.} \&
  \bibinfo{author}{Rao, M.~V.}
\newblock \bibinfo{journal}{\bibinfo{title}{Nanowire-{Based} {Sensor} {Array}
  for {Detection} of {Cross}-{Sensitive} {Gases} {Using} {PCA} and {Machine}
  {Learning} {Algorithms}}}.
\newblock {\emph{\JournalTitle{IEEE Sensors Journal}}}
  \textbf{\bibinfo{volume}{20}}, \bibinfo{pages}{6020--6028},
  \doiprefix\url{10.1109/JSEN.2020.2972542} (\bibinfo{year}{2020}).
\newblock \bibinfo{note}{Conference Name: IEEE Sensors Journal}.

\bibitem{thorson_using_2019}
\bibinfo{author}{Thorson, J.}, \bibinfo{author}{Collier-Oxandale, A.} \&
  \bibinfo{author}{Hannigan, M.}
\newblock \bibinfo{journal}{\bibinfo{title}{Using {A} {Low}-{Cost} {Sensor}
  {Array} and {Machine} {Learning} {Techniques} to {Detect} {Complex}
  {Pollutant} {Mixtures} and {Identify} {Likely} {Sources}}}.
\newblock {\emph{\JournalTitle{Sensors}}} \textbf{\bibinfo{volume}{19}},
  \bibinfo{pages}{3723}, \doiprefix\url{10.3390/s19173723}
  (\bibinfo{year}{2019}).
\newblock \bibinfo{note}{Number: 17 Publisher: Multidisciplinary Digital
  Publishing Institute}.

\bibitem{guo_smartphone-based_2021}
\bibinfo{author}{Guo, X.} \emph{et~al.}
\newblock \bibinfo{journal}{\bibinfo{title}{Smartphone-based {DNA} diagnostics
  for malaria detection using deep learning for local decision support and
  blockchain technology for security}}.
\newblock {\emph{\JournalTitle{Nature Electronics}}}
  \textbf{\bibinfo{volume}{4}}, \bibinfo{pages}{615--624},
  \doiprefix\url{10.1038/s41928-021-00612-x} (\bibinfo{year}{2021}).
\newblock \bibinfo{note}{Number: 8 Publisher: Nature Publishing Group}.

\bibitem{tsakanikas_machine_2020}
\bibinfo{author}{Tsakanikas, P.}, \bibinfo{author}{Karnavas, A.},
  \bibinfo{author}{Panagou, E.~Z.} \& \bibinfo{author}{Nychas, G.-J.}
\newblock \bibinfo{journal}{\bibinfo{title}{A machine learning workflow for raw
  food spectroscopic classification in a future industry}}.
\newblock {\emph{\JournalTitle{Scientific Reports}}}
  \textbf{\bibinfo{volume}{10}}, \bibinfo{pages}{11212},
  \doiprefix\url{10.1038/s41598-020-68156-2} (\bibinfo{year}{2020}).
\newblock \bibinfo{note}{Number: 1 Publisher: Nature Publishing Group}.

\bibitem{arduino_datasheet_nodate}
\bibinfo{author}{Arduino}.
\newblock \bibinfo{title}{Datasheet {Mega2650}}.

\bibitem{ika_datasheet_nodate}
\bibinfo{author}{IKA}.
\newblock \bibinfo{title}{Datasheet {CMAG}}.

\bibitem{li_application_2022}
\bibinfo{author}{Li, Y.}, \bibinfo{author}{Ma, T.} \& \bibinfo{author}{Wang,
  Y.}
\newblock \bibinfo{title}{Application {Status} of {Artificial} {Neural}
  {Network} {Technology} in {Clinical} {Pharmacy}}.
\newblock In \bibinfo{editor}{Sugumaran, V.}, \bibinfo{editor}{Sreedevi, A.~G.}
  \& \bibinfo{editor}{Xu, Z.} (eds.) \emph{\bibinfo{booktitle}{Application of
  {Intelligent} {Systems} in {Multi}-modal {Information} {Analytics}}}, Lecture
  {Notes} on {Data} {Engineering} and {Communications} {Technologies},
  \bibinfo{pages}{822--828}, \doiprefix\url{10.1007/978-3-031-05484-6_107}
  (\bibinfo{publisher}{Springer International Publishing},
  \bibinfo{address}{Cham}, \bibinfo{year}{2022}).

\bibitem{pattanayak_pro_nodate}
\bibinfo{author}{Pattanayak, S.}
\newblock \emph{\bibinfo{title}{Pro {Deep} {Learning} with {Tensorflow} : {A}
  {Mathematical} {Approach} to {Advanced} {Artificiall} {Intelligence} in
  {Python}}}.

\bibitem{chollet_deep_2021}
\bibinfo{author}{Chollet, F.}
\newblock \emph{\bibinfo{title}{Deep {Learning} with {Python}}}
  (\bibinfo{year}{2021}), \bibinfo{edition}{2} edn.

\bibitem{patel_drawn--skin_2022}
\bibinfo{author}{Patel, S.} \emph{et~al.}
\newblock \bibinfo{journal}{\bibinfo{title}{Drawn-on-{Skin} {Sensors} from
  {Fully} {Biocompatible} {Inks} toward {High}-{Quality} {Electrophysiology}}}.
\newblock {\emph{\JournalTitle{Small}}} \textbf{\bibinfo{volume}{18}},
  \bibinfo{pages}{2107099}, \doiprefix\url{10.1002/smll.202107099}
  (\bibinfo{year}{2022}).
\newblock \bibinfo{note}{\_eprint:
  https://onlinelibrary.wiley.com/doi/pdf/10.1002/smll.202107099}.

\end{thebibliography}
\section*{Acknowledgement}
This work was supported by Natural Sciences and Engineering Research Council of Canada (NSERC).
\section*{Author contributions statement}

S.G.C and J.P conceived the experimental approach.  J.P conducted the experiments, analysed the results and wrote the first draft of the manuscript. S.G.C and F.V reviewed and corrected the manuscript.

\section*{Competing interests}

The authors declare no competing interests.
\end{document}


\flushbottom
\maketitle

\thispagestyle{empty}
\newpage

\section*{Materials}
\subsection*{Sensors}
\noindent The DS18B20 IC-based sensor communicates over a 1-Wire bus, which implies it only needs one data line for communication with the microprocessor. Its operating range is between -55$^{\circ}$C and 125$^{\circ}$C. Its accuracy varies between 0.5-2.0$^{\circ}$C, the best obtained for temperatures in a -10 to 85$^{\circ}$C range. The NXRT15WF104FA1B040 sensor is a NTC thermistor with a 100k$\Omega$ resistance at 25$^{\circ}$C. Its operating range is between -40$^{\circ}$C and 125$^{\circ}$C. Its precision varies but stays around 2.0$^{\circ}$C in this range. 

\section*{Methods}
\subsection*{Neural Network}

As mentionned in the paper, we tried different parameters for our model in order to get an optimized version. It was clear that the hyperbolic tangent activation function for both layers worked the best. As for the number of epochs, as depicted by Figure 5 in text, the loss is already stable after 100 epochs. Therefore, adding epochs only results in overfitting after approximately 320 epochs. For 600 epochs, we get a 3.47x10$^{-2}$ loss on the test set, which is roughly a factor two hundred (200$\times$) larger than our best MSE result. The step in figure below shows the overfitting.

\begin{figure}[h!]
\begin{subfigure}{.5\textwidth}
  \centering
  \includegraphics[width=.9\linewidth]{figures/logscale.png}  
  \caption{}
  \label{fig:sub-first}
\end{subfigure}
\begin{subfigure}{.5\textwidth}
  \centering
  \includegraphics[width=.9\linewidth]{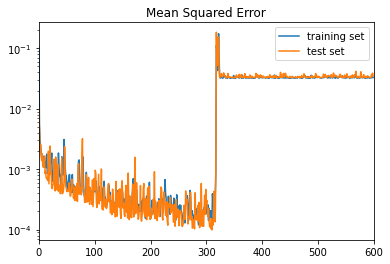}  
  \caption{}
  \label{fig:sub-second}
\end{subfigure}
\caption{(a) MSE loss function on logarithmic scale of our model (b) MSE loss function on logarithmic scale of our model for 600 epochs}
        \label{fig:three graphs}
\end{figure}

\noindent Furthermore, we tried adding an extra hidden layer to our model to study the effect. Adding a 12-neuron layer resulted in a 2.27x10$^{-4}$ loss for the test set, which is roughly twice larger than our best MSE result.

\begin{figure}[h!]
\begin{subfigure}{.5\textwidth}
  \centering
  \includegraphics[width=.9\linewidth]{figures/logscale.png}  
  \caption{}
  \label{fig:sub-first}
\end{subfigure}
\begin{subfigure}{.5\textwidth}
  \centering
  \includegraphics[width=.9\linewidth]{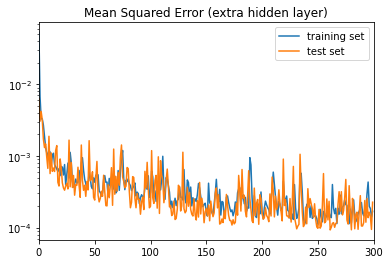}  
  \caption{}
  \label{fig:sub-second}
\end{subfigure}
\caption{(a) MSE loss function on logarithmic scale of our model (b) MSE loss function on logarithmic scale of our model with extra hidden layer}
        \label{fig:three graphs}
\end{figure}

\section*{Results}

For the paper, we trained the model with both types of sensors jointly. Here are the prediction results for the same model, but trained only with the digital readings (yellow) and the analog readings (blue). Their respective prediction are also on the graph.

\begin{figure}[h]

 \centering
    \includegraphics[width=0.75\textwidth]{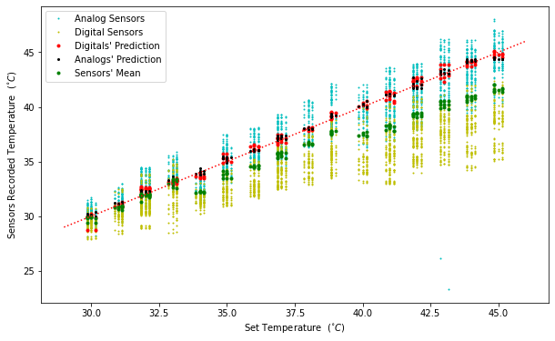}
    \caption{Independent model's prediction according to temperature sensors' type}
    \label{fig:my_label}
\end{figure}